\documentclass[journal]{IEEEtran}
\ifCLASSINFOpdf
\else
\fi
\usepackage{graphicx} 
\usepackage{epstopdf}
\usepackage{amsmath}
\usepackage{bm}
\usepackage{caption}
\usepackage{subfigure}
\usepackage{array}
\usepackage{amsfonts}
\usepackage{multirow}
\usepackage{multicol}
\usepackage{arydshln}

\begin{document}
%
\title{A Robust Imbalanced SAR Image Change Detection Approach Based on Deep Difference Image and PCANet}

%
\author{Xinzheng~Zhang,
        ~Hang~Su,~Ce~Zhang,~Peter~M.~Atkinson,~Xiaoheng~Tan,~Xiaoping~Zeng~and ~Xin~Jian
\thanks{This work was supported in part by the National Science Foundation of China under Grant Number 61301224, in part by the Basic and Advanced Research Project in Chongqing under Grant Number cstc2017jcyjA1378 and cstc2016jcyjA0134. (Corresponding author: Xinzheng Zhang.)}
\thanks{X. Zhang, H. Su, X. Tanm, X.Zeng and X.Jian are with the College of Microelectronic and Communication Engineering, Chongqing University, Chongqing 400044, China (e-mail: zhangxinzheng@cqu.edu.cn).}
\thanks{X. Zhang,X. Tan, X.Zeng and X.Jian are with the Chongqing Key Laboratory of Space Information Network and Intelligent Information Fusion, Chongqing, 400044, China.}
\thanks{C. zhang and Peter M. Atkinson are with the Lancaster Environment Centre, Lancaster University, Lancaster, LA1 4YQ, United Kingdom; C. zhang, and Peter M. Atkinson are also with the UK Centre for Ecology \& Hydrology, Library Avenue, Lancaster, LA1 4AP, United Kingdom }.}
\maketitle
\begin{abstract}
In this research, a novel robust change detection approach is presented for imbalanced multi-temporal synthetic aperture radar (SAR) image based on deep learning. Our main contribution is to develop a novel method for generating difference image and a parallel fuzzy c-means (FCM) clustering method. The main steps of our proposed approach are as follows: 1) Inspired by convolution and pooling in deep learning, a deep difference image (DDI) is obtained based on parameterized pooling leading to better speckle suppression and feature enhancement than traditional difference images. 2) Two different parameter $\bf Sigmoid$ nonlinear mapping are applied to the DDI to get two mapped DDIs. Parallel FCM are utilized on these two mapped DDIs to obtain three types of pseudo-label pixels, namely, changed pixels, unchanged pixels, and intermediate pixels. 3) A PCANet with support vector machine (SVM) are trained to classify intermediate pixels to be changed or unchanged. Three imbalanced multi-temporal SAR image sets are used for change detection experiments. The experimental results demonstrate that the proposed approach is effective and robust for imbalanced SAR data, and achieve up to 99.52\% change detection accuracy superior to most state-of-the-art methods.
\end{abstract}

%
\IEEEpeerreviewmaketitle

\section{INTRODUCTION}
%
%
%
%

\IEEEPARstart{S}{AR} image change detection technology is widely used in earth observation tasks such as environmental protection, urban research, and forest resource monitoring [1]. Compared with optical images, multi-temporal SAR images are usually contaminated by inherent speckle noise, which impacts much negative effects on change detection algorithms [2]-[5].

Due to the extreme shortage of multi-temporal SAR images with ground truth, supervised methods are limited in SAR image change detection. At present, unsupervised methods are widely studied and applied in this field. The main steps of unsupervised methods usually include: 1) Preprocessing; 2) difference image (DI) generation; 3) classification. A common method of generating DI is to calculate the log-ratio of the two SAR images pixel by pixel, the obvious disadvantage of which is the poor noise robustness. In classification, unsupervised clustering algorithms are applied at early stages. For example, $k$-means clustering [6], multiple kernel $k$-means clustering [7], and fuzzy $c$-means clustering [8] have been extensively studied to classify pixels or produce pseudo-label training samples to subsequent classifiers from a DI. However, these existing clustering algorithms have significant disadvantages. On one hand, these distance-based clustering algorithms are extremely sensitive to speckle noise, which often leads to treat changes caused by speckles as real terrain objects changes. On the other hand, most of the existing clustering algorithms assume that the changed and unchanged classes are balanced. In many cases, the pixels of changed class are far less than the those of unchanged class, which is typical imbalanced. Traditional clustering methods will cause excessive false alarms when faced with imbalanced data.  

In recent years, deep learning exhibits excellent performance in pattern recognition. Many researchers have introduced deep learning into SAR image change detection and achieved superior performance. Gao \emph {et al}. use cluster-based PCANet [9] and CWNN [10] to exploit the key information, respectively. In [3], Geng \emph{et al}. proposed a saliency-guided deep neural network for SAR image change detection. Li \emph {et al}. proposed a SAR image change detection method based on convolutional neural network (CNN) [11]. Recently, researchers have noticed the imbalance in SAR image change detection. Wang \emph{et al}. proposed an imbalanced learning method by morphologically supervised PCANet [12]. To our best knowledge, there are still few studies on the imbalanced SAR images change detection, it is a fact that deep learning methods have extremely high requirements on the quality of training samples. However, the problem is that, with strong speckle noise and imbalanced data, it is hard to obtain high reliable pseudo-label training samples for deep learning models. Apparently, high quality DI and effective clustering methods are benefit for this issue. 

In this research, a novel approach based on deep learning is proposed for imbalanced SAR image change detection. Especially, a novel DI generation method to get deep difference image (DDI) and a parallel FCM clustering (PFCMC) method are developed in the proposed approach. Our approach integrates DDI, PFCMC and PCANet to implement imbalance multi-temporal SAR image with strong speckle noise change detection, which is abbreviated as DP\_PCANet. Our main contributions are as follows:

\setlength{\hangindent}{0.8cm}1)  Based on the ideas of convolution and pooling in deep learning, we developed  weighted pooling and cumulative weighted pooling. This method of generating DDI can effectively suppress various speckle noise and enhance terrain changes than traditional method.

\setlength{\hangindent}{0.8cm}2)  PFCMC method is developed for imbalanced SAR image based on combining nonlinear $Sigmoid$ mapping, Gabor wavelets and basic FCM, it can provide more reliable pseudo-labels training samples in the case of significantly imbalanced SAR images.

\setlength{\hangindent}{0.8cm}3)  In PCANet classification, over-sampling and under-sampling are employed to reduce the negative impact of imbalanced data.

\vspace{0.1cm}
\setlength{\abovecaptionskip}{0.1cm}
\setlength{\belowcaptionskip}{-0.4cm}
\begin{figure}[t]
\centering
\includegraphics[height=10cm,width=7.8cm]{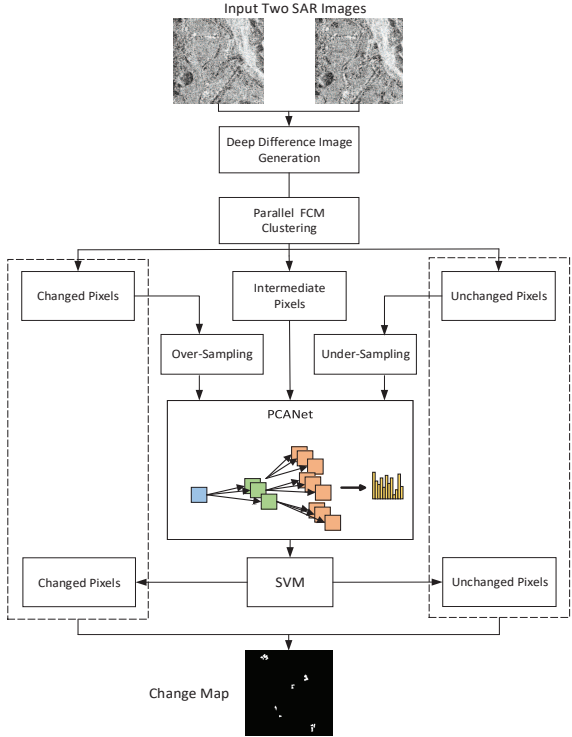}
\captionsetup{font={scriptsize},name=Fig.,labelsep=period}
\caption{The flowchart of the proposed change detection approach.}
\label{fig_sim}
\end{figure}
The remainder of this letter is organized as follows. Section II  describes the proposed approach. Experimental results, comparison and analysis are in Section III. Finally, the conclusion is given in Section IV.
\section{METHODOLOGY}
In this letter, the proposed approach mainly includes three parts: 1) generating a DDI; 2) parallel FCM clustering; 3) a PCANet with SVM are designed to classify pixels based on pseudo-label training samples. The flowchart of the proposed approach is illustrated in Fig. 1.
\subsection{Deep Difference Image Gerneration }
In deep CNN, average-pooling and maximum-pooling are important processes to extract deep features. Although average-pooling is effective in suppressing noise, it can also easily result in excessive loss of details information. As for maximum-pooling, it can effectively enhance image features, however, it is more sensitive to noise. Motivated by the above ideas, we have proposed a weighted-pooling method for generating a DDI based on the distance measure of the local window.

If the local window is size of $k\times k$, where $k$ is odd , the weighted-pooling kernel ${\bf W}^k$ is a matrix with the same size as shown in Equation (1). Each element value is determined by the distance between the current position and the matrix center, which is calculated by Equation (2).    
\begin{equation}
{\bf W}^k = \left(
\begin{array}{cccc}
w_{11} & w_{12} & \ldots & w_{1k}\\
w_{21} & w_{22} & \ldots & w_{2k}\\
\vdots & \vdots & \ddots & \vdots\\
w_{k1} & w_{k2} & \ldots & w_{kk}\\ 
\end{array} \right)
\end{equation}
\begin{equation}
w_{ij} = \frac{1}{k^2\sqrt{(\frac{k+1}{2}-i)^2+(\frac{k+1}{2}-j)^2}}; \\  i,j = 1,2,...,k
\end{equation}

As a special case, the value of the element at the center of the matrix is defined as $w_{(k+1)/2,(k+1)/2}={2}/{k^2}$. After weighted-pooling convolution, each pixel in the weighed-pooling image can be represented by Equation (3), where $I_{n,m}$ is pixel in the original image
\begin{equation}
I_{nm}^{wp} = \frac{1}{k^2}\times{\sum_{i = 1}^{k}\sum_{j = 1}^{k}w_{ij}\times I_{n+i-(k+1)/2,n+j-(k+1)/2}}
\end{equation}

Given two multi-temporal SAR images ${\bf I}_1$ and ${\bf I}_2$, where ${\bf I}_1,{\bf I}_2 \in \mathbb{R}^{N\times M}$, convolution of each SAR image using the same weighted pooling kernel to obtain two pooled images ${\bf I}_1^{wp}(k) = {\bf I}_1 \ast {\bf W}^k$ and ${\bf I}_2^{wp}(k) = {\bf I}_2 \ast {\bf {W}}^k$, where $\ast$ denote 2-D convolution operation. Then, the log-ratio image ${\bf I}_d$ is calculated as:
\begin{equation}
{\bf I}_d = |log({\bf I}_2^{wp}/{\bf I}_1^{wp})|
\end{equation}

The log-ratio operator has weak anti-noise performance, there is still a certain degree of speckle noise in ${\bf I}_d$. Based on the sparse distribution of noise outliers and the aggregated distribution of real change in log-ratio image, we apply weighted-pooling operation with different local window size to  ${\bf I}_d$ for $T$ times, obtaining $T$ pooled images. Then, all $T$ pooled images is accumulated to generate the DDI denoted by ${\bf I}_{DDI}$:
\begin{equation}
{\bf I}_{DDI}= \frac{1}{T} \sum_{t=1}^{T}\frac{{\bf I}_d^{wp}(2t-1)}{\bar w}
\end{equation}
where ${\bar w}=\frac {1}{(2t-1)^2}\sum_{i=1}^{2t-1}\sum_{j=1}^{2t-1}w_{ij}$ and ${\bf I}_d^{wp}(2t-1)$ denotes a pooled image of ${\bf I}_d $ after weighted-pooling with parameter $ k=2t-1$.The process of generating the DDI is exhibited in Fig. 2.
\vspace{0.1cm}
\setlength{\abovecaptionskip}{0.1cm}
\setlength{\belowcaptionskip}{-0.4cm}
\begin{figure}[!t]
\centering
\includegraphics[height=3.2 cm,width=8.8cm]{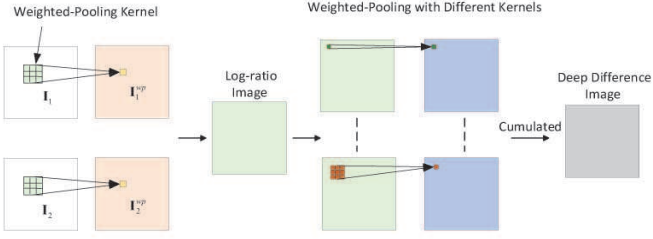}
\captionsetup{font={scriptsize},name=Fig.,labelsep=period}
\caption{The process of generating a deep difference image.}
\label{fig_sim}
\end{figure}

\subsection{Parallel FCM Clustering}
In SAR image change detection, the number of changed pixels is often much less than that of unchanged pixels, The traditional clustering methods have poor clustering effect in case of imbalanced data.We develop parallel FCM clustering based on $Sigmoid$ mapping method to generate high quality pseudo labels. In order to enhance the difference of pixel categories and effectively reduce the impact of sample imbalance characteristics, the PFCM method first employs two sigmoid functions to map DDI to obtain two mapped images. The Sigmoid mapping function is with two parameters $\gamma$ and $\mu$, and a variable $x$ is mapped by (6).
\begin{equation}
\emph sig(x;\gamma,\mu)= \frac{1}{1+e^{-\gamma(x+\mu)}}
\end{equation}

The Gabor wavelet transform is used to extract features from the two mapped images to obtain two sets of pixel-level feature vectors ${\bf X}^1$ and ${\bf X}^2$ respectively, details of Gabor feature extraction can be referred in [9]. At last, FCM and simple coding method are used to get the final clustering results. The detailed descriptions of the parallel FCM algorithm are as follows:

\setlength{\hangindent}{0.85cm}1) {\emph {Input}}: Deep difference image ${\bf I}_{DDI}$.

\setlength{\hangindent}{0.85cm}2) {\emph {Step1}}: Obtain ${\bf I}_{DDI}^C$ by normalizing and centralizing ${\bf I}_{DDI}$, apply two $Sigmoid$ mapping to ${\bf I}_{DDI}^C$ with two different parameter sets pixel-by-pixel to get ${\bf I}_1^M = {\emph sig}({\bf I}_{DDI};\gamma_1,\mu_1)$  and  ${\bf I}_2^M = {\emph sig}({\bf I}_{DDI};\gamma_2,\mu_2)$.

\setlength{\hangindent}{0.85cm}3) {\emph{Step2}}: Perform Gabor feature extraction to ${\bf I}_1^M$ and ${\bf I}_2^M$ respectively. Two Gabor feature vector sets are obtained, which are denoted as ${\bf X}^1 = [{\bf x}_1^1, {\bf x}_2^1,...,{\bf x}_{NM}^1]$ and ${\bf X}^2 = [{\bf x}_1^2, {\bf x}_2^2,...,{\bf x}_{NM}^2]$, where ${\bf x}_s^p (s=1,2,...,NM; p=1,2 )$ represents a Gabor feature vector.

\setlength{\hangindent}{0.85cm}4) {\emph{Step3}}: FCM is utilized to perform two class clustering on ${\bf X}^1$ and ${\bf X}^2$ to get two label sets ${\bf Y}^1 = [y_1^1,y_2^1,..., y_{NM}^1]$ and ${\bf Y}^2 = [y_1^2,y_2^2,..., y_{NM}^2]$ respectively, where $ y_s^p (s=1,2,...,NM; p=1,2 )$ represents a label corresponding to a Gabor feature vector. The value of   $y_s^p$ is 0 or 1. The simple averaging operation is used to encode the two label sets, obtaining the final label set ${\bf Y} = [y_1,y_2,..., y_{NM}]$ , where $y_s = (y_s^1+y_s^2)/2$.

\setlength{\hangindent}{0.85cm}5) {\emph {Step4}}: If $y_s = 1$, assign the corresponding pixel to the changed class $\omega_c$, if $y_s = 0$, assign the corresponding pixel to the unchanged class $\omega_{uc}$, the others are assigned to the intermediate class $\omega_i$. 

\setlength{\hangindent}{0.85cm}6) {\emph {Output}}: The clustering result can be denoted by an image with labels \{$\omega_c$,  $\omega_{uc}$,  $\omega_i$\}. 

The pixels belonging to $\omega_c$ have the high probability to be changed, while the pixels belonging to $\omega_{uc}$ have the high probability to be unchanged. These two kinds of pixels can be chosen as samples for training PCANet model to classify those pixels belonging to $\omega_i$. 

\subsection{PCANet Classification Model}
PCANet is a kind of deep learning model with strong robustness to noise [13]. In this letter, a PCANet with the similar two stages structure in [9] and linear SVM are adopted as the classification model.

Patches of size $\lambda \times \lambda$ are generated pixel by pixel on two weighted-pooled SAR images ${\bf I}_1^{wp}$ and ${\bf I}_2^{wp}$. The corresponding patches are connected to form a new patch set denoted as ${\bf P}^1 = [{\bf P}_1, {\bf P}_2,...,{\bf P}_{NM}]$ , where ${\bf P}_e \in \mathbb{R}^{2\lambda \times \lambda}, e = 1,2,...,NM$. Over-sampling the pixels of $\omega_c$ and down-sampling the pixels of $\omega_{uc}$ to maintain a certain balance between the sample numbers. Randomly select $S$ image-patch samples ${\bf P}^1 = [{\bf P}_1, {\bf P}_2,...,{\bf P}_S]$  and vectorize them to obtain feature vectors, then subtract vector mean from each a vector and combine them to a new feature matrix ${\bf {\bar P}}^1 = [{\bf {\bar P}}_1, {\bf {\bar P}}_2,...,{\bf {\bar P}}_S] \in \mathbb{R}^{2\lambda ^2 \times S}$. The expression for calculating the PCA filter is as:
\begin{equation}
{\bf Q}_{l_1}^1 = mat(q_l({\bf P}_t{\bf P}_t^{\bf T})) \in \mathbb{R}^{2\lambda ^2 \times 2\lambda ^2}
\end{equation}
where $mat(\bf v)$ is a function that maps ${\bf v} \in \mathbb{R}^{4\lambda ^4}$ to a matrix ${\bf Q} \in \mathbb{R}^{2\lambda ^2 \times 2\lambda ^2}$, and $q_l({\bf P}_t{\bf P}_t^{\bf T})$ means the $l_1$-\emph {th} principle eigenvector of ${\bf P}_t{\bf P}_t^{\bf T}$. Then the $l_1$-\emph {th} filter output of the first stage in PCANet is:
\begin{equation}
{\bf R}_s^{l_1} = {\bf P}_s \ast {\bf R}_{l_1}^1  \  \  s=1,2,...,S
\end{equation}
where $\ast$ denotes 3-D convolution. 

The first layer of PCANet calculation is completed to obtain ${\bf R}_s=[{\bf R}_s^1,{\bf R}_s^2,...,{\bf R}_s^{L_1}],{\bf R}_s^{l_1}\in \mathbb{R}^{2\lambda \times \lambda}$. The process of the second stage is similar to the first stage, the same method is applied to  each  ${\bf R}_s^{l_1}$ to get the second stage output  ${\bf Z}_s^{l_1}= \{ {\bf Z}_s^{{l_1}{l_2}}\}_{{l_2} =1}^{L_2}$ where ${\bf Z}_s^{{l_1}{l_2}}\in \mathbb{R}^{2\lambda \times \lambda}$.
\begin{equation}
{\bf Z}_s^{l_1l_2} = {\bf R}_s^{l_1} \ast {\bf Q}_{l_1}^2  \  \  s=1,2,...,S; l_2 = 1,2,...,L_2
\end{equation}

We binarize outputs to obtain ${\bf Z}_s^{l_1}= \{ H({\bf Z}_s^{{l_1}{l_2}})\}_{{l_2} =1}^{L_2}$, where $H(\cdot)$ is a Heaviside step function, whose value is one for positive and zero otherwise. Around each pixel of ${\bf R}_s^{l_1}$, the vector of  $L_2$ binary bits can be converted into an integer value, the conversion formula is as follows:
\begin{equation}
T_s^{l_1} = \sum_{{l_2}=1}^{L_2}2^{{l_2}-1}H({\bf Z}_s^{{l_1}{l_2}})
\end{equation}

The single integer-value image $T^{l_1}$ is obtained, and each pixel $T_s^{l_1}$ is an integer in the range $[0,2^{L_2-1}]$. We further transform  $T_s^{l_1}$ into a histogram denoted by $hist(T_s^{l_1})$. Then the PCANet feature of the image-patch ${\bf P}_s$  can be defined as 
\begin{equation}
{\bf f}_s=[hist(T_s^1),hist(T_s^2),...,hist(T_s^{l_1})]
\end{equation}

Every image-patch ${\bf P}_s$ is processed by the above method to get PCANet feature, and the extracted features and clustering labels are fed into linear SVM to train model. Use the model to further classify the pixels belonging to $\omega_i$ into $\omega_c$ and $\omega_{uc}$. Reconstruct the labels by position to get the final changed map. 

\vspace{0.1cm}
\setlength{\abovecaptionskip}{0.1cm}
\setlength{\belowcaptionskip}{-0.4cm}
\begin{figure*}[t]
\centering
\includegraphics[height=5cm,width=18.2cm]{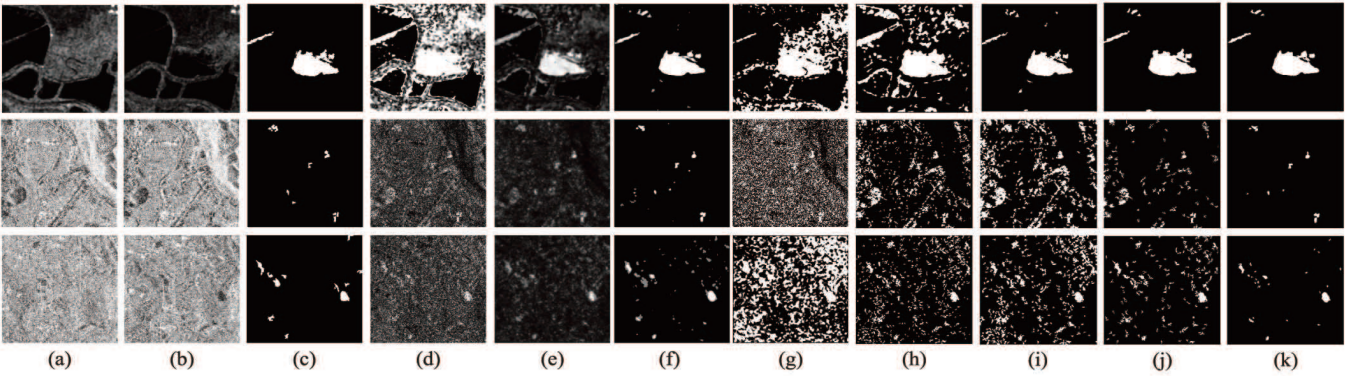}
\captionsetup{font={scriptsize},name=Fig.,labelsep=period}
\caption{Visualize results.The first line is for dataset A, the second line is for dataset B, and the third line is for dataset C. (a) and (b) show two original SAR images, respectively; (c) The Ground truth; (d) The DI  generated by the traditional log-ratio operator; (e) The DDI generated by our proposed method; (f) Parallel FCM clustering results; (g) Results of PCA\_kmeans; (h) Results of NR\_ELM; (i) Results of Gabor\_PCANet ; (j) Results of CWNN; (k) Results of our proposed approach. }
\label{fig_sim}
\end{figure*}

\section{EXPERIMENTAL RESULTS AND ANALYSES}
\subsection{Experimental Setup}
In order to evaluate the performance of the proposed approach, we apply three real SAR datasets. The first is called the A dataset, which presents a section of two SAR images acquired by ERS-2 SAR sensor over the city of San Francisco. Each of these images is size of $256\times 256$. The second dataset and the third dataset are named the B and C dataset respectively, which presents two SAR images captured by the COSMO-Skymed SAR sensor with a size of $400\times 400$. The SAR images in B and C contain relatively strong speckle noise, and the areas of changed and unchanged in these SAR images are extremely imbalanced.  

The following indicators are employed to evaluate the proposed approach: false positives (\emph {FP}), false negatives (\emph {FN}), percentage correct classification (\emph {PCC}) and Kappa coefficient (\emph {KC}). The \emph {FP} denotes the number of unchanged pixels that are false detected, while \emph {FN} refers the number of real changed pixels are detected as unchanged. True negatives (\emph {TN}) indicates the unchanged pixels that are correctly detected, and true positives (\emph {TP}) is the correct detection of changed pixels. Let \emph {Nc} and \emph {Nu} refer the number of truly changed and unchanged pixels of the ground truth respectively. We employ $\emph {IR}=\emph {Nc}/\emph{Nu}$ to measure the imbalance between the changed class and unchanged class. Overall errors (\emph {OE}) $\emph {OE}=\emph {FN}+\emph {FP}$ is utilized to indicate the number of pixels for all detected errors. The PCC refers the ratio of correctly classified pixels to all pixels, which is expressed as $\emph {PCC}={(\emph {TP}+\emph {TN})}/{(\emph {TP}+\emph {TN}+\emph {FP}+\emph {FN})}$.
The \emph {KC} is considered as an essential criterion, which is for consistency check. It is calculated as:
\begin{equation}
\emph {PRE} =\frac{(\emph {TP}+\emph {FP})\times (\emph {TP}+\emph {FN})+(\emph {FN}+\emph {TN})\times (\emph {FP}+\emph {TN})}{{\emph {TP}+\emph {TN}+\emph {FP}+\emph {FN}}^2}
\end{equation}
\begin{equation}
\emph {KC} =\frac{\emph {PCC}-\emph {PRE}}{1-\emph {PRE}}
\end{equation}

For comparison, four state-of-the-art approaches are utilized on the above three datasets. These approaches are PCA\_kmeans [14] , NR\_ELM [15], Gabor\_PCANet [9] and CWNN[10]. Moreover, the parameters of these methods are set with reference to the corresponding papers.

In the proposed DP\_PCANet approach, we set $k=3$ and $\gamma=7$, respectively. In order to reduce the number of hyperparameters, we define the relationship and constraint between the two sets of the center biases $b$ in the $sig$ function:$b=(\mu_1+\mu_2)/2 \   \   \emph {s.t.}|\mu_1-\mu_2|=0.12  $.

As mentioned in [9], the parameter of PCANet include image-patch size $\lambda$, the filter numbers $L_1$ and $L_2$, and the training samples $S$. We set $\lambda =5$ and $L_1=L_2=8$ the $S = 0.2\times NM$. At the stage of generating DDI and parallel clustering, we set $T=7,b=0.1$ for A; $T=11,b=0.17$ for B; $T=11,b=-0.04$ for C.

\subsection{Analysis of Result}
Comparing (d) and (e) in Fig. 3, it can be found that The proposed method of generating DDI can effectively suppress speckle noise and enhance the characteristics of the real changed area for all three datasets. This is mainly because the weighted pooling has noise reduction on the original image, in addition, the image features are effectively retained. Then, the cumulated weighted-pooling can further weaken the sparse high-value noise on the log-ratio image, and further strengthen real changes. The DDI is significantly better than traditional DI visually.

The imbalanced ratio was calculated: $IR_A=0.077$ ,$IR_B=0.0094$ and $IR_C=0.0221$. As can be seen from (f) in Fig. 3, the PFCMC method can effectively adapt to images of different imbalanced ratio, and clustering results image is satisfied for all three datasets. 

In Table I, we exhibit the values of the above-mentioned indicators to evaluate these change detection methods. The \emph{PCC} and \emph {KC} of DP\_PCANet are 99.08\% and 0.9296, respectively, which are slightly higher than 98.90\% and 0.9164 of Gabor\_PCANet, 98.94\% and 0.9231 of CWNN. It demonstrates that the change detection result of DP\_PCANet is almost identical to the Ground truth. For the dataset B with strong speckle noise, the \emph {PCC} of DP\_PCANet reaches 99.52\%, the \emph {KC} reaches 0.7177. For C dataset, the \emph {PCC} of DP\_PCANet reaches 98.65\% and the \emph {KC} is 0.6081. The indicators of our proposed algorithm are higher than other methods on three datasets. Comparing the change detection results of these five approaches in Fig. 3, we draw the conclusion that proposed approach is superior to other methods. Above results illustrates that our algorithm's robustness and generalization are significantly better than other methods on the three sets of data, and have extremely strong detection performance. 

The degree of imbalance in the three datasets is quite different, the imbalance of dataset B is very serious, especially. Various imbalance in these datasets has a strong negative effect on the clustering accuracy and the performance of the deep learning classification model. Due to imbalanced characteristic, the other four methods identify large number of unchanged pixels as changed pixels, leading to great false alarms. Compared with other methods, the \emph {FP} of our proposed approach are 221, 266, and 585 on the three sets of data. DP\_PCANet has the lowest \emph {FP} and maintains the lowest \emph {OE}, while keeping the highest \emph{PCC} and \emph{KC}. Experimental results demonstrate that DP\_PCANet has strong adaptive ability and generalization performance on imbalanced multi-temporal SAR image change detection tasks.
\begin{table}[tp]  
\centering
\caption{CHANGE DETECTION RESULTS OF DIFFERENT METHODS ON THREE DATASETS.}  
\label{tab:methodcompare}
\begin{tabular}{p{1.8cm}<{\centering}|p{0.9cm}<{\centering} p{0.9cm}<{\centering} p{0.9cm}<{\centering}p{0.9cm}<{\centering}p{0.9cm}<{\centering}}
\hline
\hline
\multirow{2}*{Model} & \multicolumn{5}{c}{Results on the A dataset}\\ 
\cline{2-6}
        &\emph {FP} &\emph {FN} &\emph {OE} &\emph {PCC} &\emph {KC}\\
\hline 
 PCA\_kmeans    &13496 &218 &13714 &0.7907 &0.3170\\
 NR\_ELM        &10137 &10 &10147	&0.8452 &0.4161\\
 Gabor\_PCANet  &311 &409 &720	&0.9890 &0.9164\\
 CWNN           &545 &150 &695	&0.9894 &0.9231\\
{\bf DP\_PCANet}    &{\bf221} &{\bf 382} &{\bf 603}	&{\bf 0.9908} &{\bf 0.9296}\\
\hline

\hline
\multirow{2}*{Model} & \multicolumn{5}{c}{Results on the B dataset}\\ 
\cline{2-6}
        &\emph {FP} &\emph {FN} &\emph {OE} &\emph {PCC} &\emph {KC}\\
\hline 
 PCA\_kmeans    &60119 &241 &60360 &0.6228 &0.0220\\
 NR\_ELM        &20721 &236 &20957	&0.8690 &0.0912\\
 Gabor\_PCANet  &23761 &131 &23892	&0.8507 &0.0862\\
 CWNN           &7194 &417 &7611	&0.9524 &0.2077\\
{\bf DP\_PCANet}    &{\bf 266} &{\bf 503} &{\bf 769}	&{\bf 0.9952} &{\bf 0.7177}\\ 
\hline

\hline
\multirow{2}*{Model} & \multicolumn{5}{c}{Results on the C dataset}\\ 
\cline{2-6}
        &\emph {FP} &\emph {FN} &\emph {OE} &\emph {PCC} &\emph {KC}\\
\hline 
PCA\_kmeans    &62095 &180 &62275 &0.6108 &0.0567\\
 NR\_ELM        &15714 &1121 &16835	&0.8948 &0.1885\\
 Gabor\_PCANet  &20860 &695 &21555	&0.8653 &0.1733\\
 CWNN           &7854 &898 &8752	&0.9453 &0.3487\\
{\bf DP\_PCANet}    &{\bf 585} &{\bf 1663} &{\bf 2248}	&{\bf 0.9860} &{\bf 0.6092}\\
\hline
\hline
\end{tabular}
\end{table}
\subsection{Analysis of Parameter b and T}
We choose a set of $T$  to explore the relationship between the cumulative times $T$ and the performance of DP\_PCANet, $KC$ is adopted as the evaluation indicator. The same way is also used to analyze the center bias $b$. The results are exhibited in Fig. 4.

From Fig. 4, it can be found that the parameter $T$ and $b$ have little effect on the dataset A, our algorithm can always maintain excellent performance for data A, the highest $KC$ is 0.9296 when $T =7$ and $b=0.1$. For dataset B and C with strong speckle noise, the result becomes better as $T$ increases. When $T$ reaches 9 and more, the $KC$ tends to be stable and maintain a highe\-level. The above experimental results prove that more cumulative times has a stronger effect on reducing speckle noise and retaining the true change. In theory, the benchmark of center bias is 0. The clustering algorithm has no preference of any kind when $b = 0$. Experimental results show the $PCC$ and $KC$ are 98.76\%, 0.5542 for data B and 98.60\%, 0.5549 for data C when $b=0$, it explains the benchmark parameters allow DP\_PCANet to show good performance in most data. there is a most suitable center biases $b$ for SAR data with different characteristics, It is worth further research to automatically select appropriate $b$.
\vspace{0.1cm}
\setlength{\abovecaptionskip}{0.1cm}
\setlength{\belowcaptionskip}{-0.4cm}
\begin{figure}[t]
\centering
\includegraphics[height=3cm,width=4.3cm]{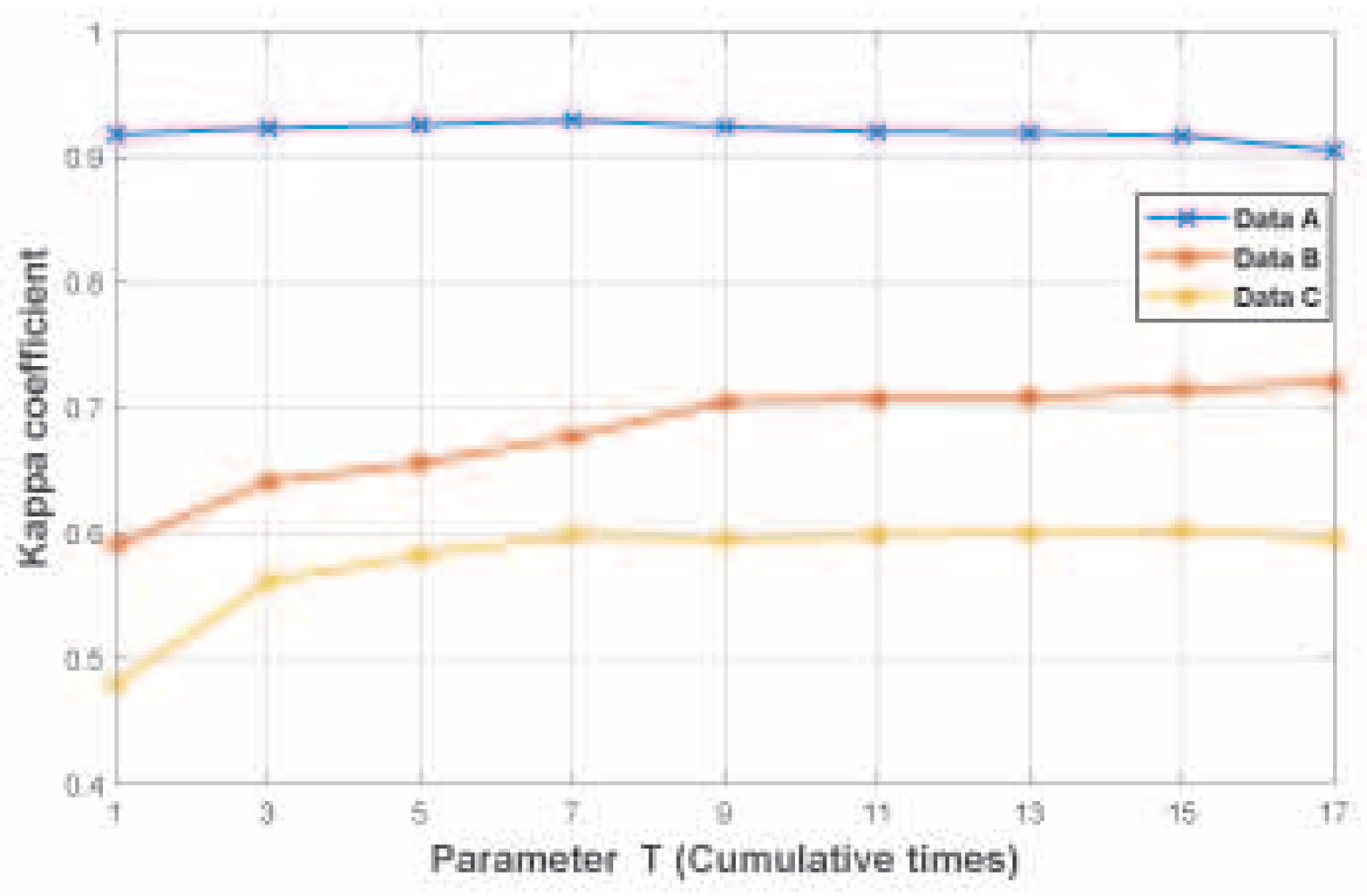}
\includegraphics[height=3cm,width=4.3cm]{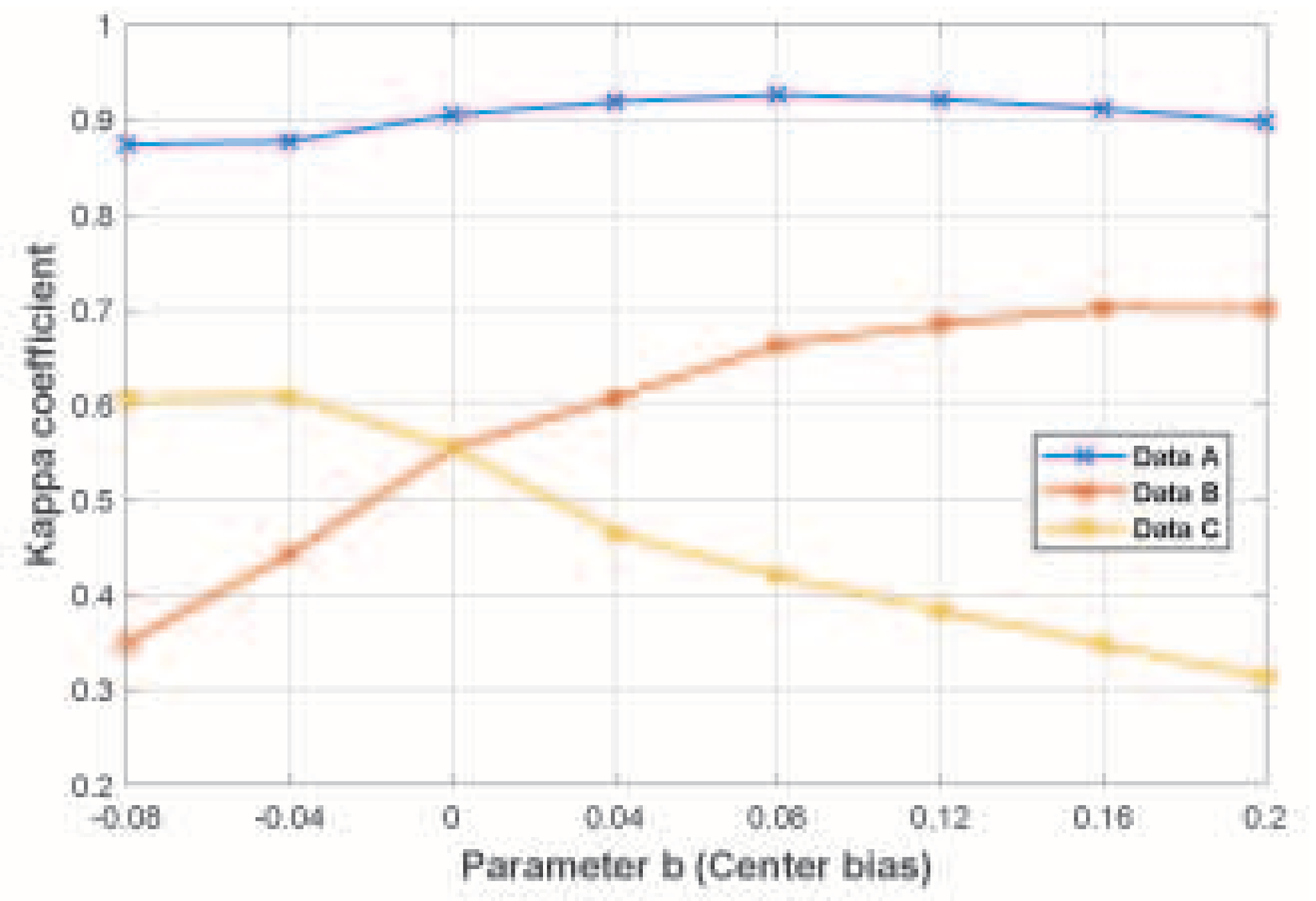}
\captionsetup{font={scriptsize},name=Fig.,labelsep=period}
\caption{Relationship between $KC$ and parameter}
\label{fig_sim}
\end{figure}
\section{Conclusion}
In this letter, we have proposed a robust  imbalanced  multi-temporal SAR change detection approach. The proposed DDI generation method can effectively reduce speckle noise and enhance features. A parallel FCM clustering method was developed to increase the gap between the change class and the unchanged class, which can obtain excellent clustering performance under imbalanced data, thereby providing highly reliable pseudo-label training samples. Over-sampling and under-sampling were employed to mitigate the imbalance effects on PCANet. Experiment results confirmed the effectiveness, generalization and robustness of the proposed approach.

\end{document}